\documentclass{article}

\PassOptionsToPackage{numbers, compress}{natbib}

     \usepackage[preprint]{neurips_2020}

\usepackage[utf8]{inputenc} 
\usepackage[T1]{fontenc}    
\usepackage{hyperref}       
\usepackage{url}            
\usepackage{multirow, booktabs}       
\usepackage{amsfonts, amsmath, bm}       
\usepackage{nicefrac}       
\usepackage{microtype}      
\usepackage[pdftex]{graphicx}
\usepackage{subcaption}
\usepackage[ruled,vlined]{algorithm2e}

\title{Stanza: A Nonlinear State Space Model for Probabilistic Inference in Non-Stationary Time Series}

\author{%
  Anna K.~Yanchenko \\
  Department of Statistical Science\\
  Duke University\\
  Durham, NC 27708 \\
  \texttt{anna.yanchenko@duke.edu} \\
   \And
   Sayan Mukherjee \\
  Department of Statistical Science\\
  Duke University\\
  Durham, NC 27708 \\
  \texttt{sayan@stat.duke.edu} \\
}

\begin{document}

\maketitle

\begin{abstract}
  Time series with long-term structure arise in a variety of contexts and capturing this temporal structure is a critical challenge in time series analysis for both inference and forecasting settings.  Traditionally, state space models have been successful in providing uncertainty estimates of trajectories in the latent space.  More recently, deep learning, attention-based approaches have achieved state of the art performance for sequence modeling, though often require large amounts of data and parameters to do so. We propose Stanza, a nonlinear, non-stationary state space model as an intermediate approach to fill the gap between traditional models and modern deep learning approaches for complex time series.  Stanza strikes a balance between competitive forecasting accuracy and probabilistic, interpretable inference for highly structured time series. In particular, Stanza achieves forecasting accuracy competitive with deep LSTMs on real-world datasets, especially for multi-step ahead forecasting. 
\end{abstract}

\section{Introduction}\label{sec:intro}

Time series with long-term structure arise in a variety of contexts and capturing this temporal structure is a critical challenge in time series analysis for both inference and forecasting. Traditionally, state space models like Kalman Filters \citep{Kalman:1960}, Dynamic Linear Models (DLMs) \citep{West-Harrison} and their numerous extensions \citep{Ljung:1979, Wan-Merwe:2000} have been utilized for probabilistic inference in time series analysis.  This general class of state space models has been very successful for modeling systems where the dynamics are approximately linear and relatively simple.  Alternatively, 
 deep learning, attention-based models \citep{NIPS2017_7181, NIPS2015_5653} have achieved state-of-the-art forecasting performance in modeling longer-term structure in time series and sequences, especially for applications relating to language. However, these methods often require large amounts of training data.  
 
 Traditional Kalman Filters and DLMs are very good for inference tasks, especially in settings where probabilistic filtering and smoothing of the state space is important.  However, Kalman Filters often are not competitive for forecasting, especially for data that does not meet the model specifications.  On the other hand, deep learning models like Long Short-Term Memory (LSTM) models \citep{LSTM:1997} achieve excellent forecasting accuracy in many settings, but cannot be used for inference tasks, as filtering and smoothing is done implicitly and deterministically in the model.  Thus, there still remains a gap between time series models designed for inference and those that achieve competitive forecasting accuracy.

We propose \textbf{Stanza}, a \underline{St}ate Sp\underline{a}ce Model with \underline{N}onlinear Activations and Dynamically Weighted Latent (\underline{Z}) St\underline{a}tes, as an intermediate approach to alleviate this trade-off between inference and forecasting. Our motivation is to fill the gap between large deep learning, attention-based models and traditional state space models. Stanza achieves forecasting accuracy comparable to deep LSTMs on three real-world datasets \textit{and} is able to perform inference for complex, nonlinear, non-stationary time series.  Furthermore, Stanza is fully probabilistic in the state space, allowing for straightforward multi-step ahead forecasts out to long time horizons and provides
forecast coverage estimates which are not possible in traditional LSTMs.  Stanza incorporates ideas from deep learning into traditional state space models, resulting in a nonlinear, non-stationary model capable of performing inference in more general and complex settings, as compared to classical state space models. We design Stanza specifically for settings where there is not enough data for a full attention-based model (i.e. the small to medium data regime) and/or where the model needs to be interpretable and uncertainty in the inferential task needs to be quantified.

The optimal setting for Stanza is quite general and occurs in many different application areas.  For example, in many medical settings, the quantity of data may be limited, yet it is critical to have an interpretable and probabilistic model for inference tasks and simple DLMs may not be rich enough to capture the complexity in the data, for example, with EEG signals \citep{West-Prado:1999}.  In retail sales settings, there may be an abundance of data, yet the goal may be to understand seasonality patterns in consumer behavior, necessitating a model with interpretable parameters that can either uncover structure in the data or specifically incorporate known structure into the model \citep{BerryWest2018DCMM}.  Finally, even in financial settings where forecasting is more important than inference, it may be the case that there is not enough data to train a full deep learning, attention-based model, yet competitive forecasting accuracy is still required (for example, in the case of the Exchange Rate data considered in \citep{Lai:2017}).  Stanza is motivated by a need for powerful inference \textit{and} forecasting. 


Our contributions are as follows:

\begin{enumerate}
\item \textbf{Stanza:} We propose Stanza as a nonlinear, non-stationary, interpretable time series model that gives full uncertainty estimates for parameters.
\item\textbf{Versatility:} We demonstrate Stanza's ability to model complex and highly structured time series and provide interpretable parameters with uncertainty estimates
for inference. We also demonstrate that Stanza  is competitive with deep LSTMs in terms of forecasting accuracy on real-world datasets, particularly for multi-step ahead forecasts.
\item\textbf{Intermediate Approach:} Stanza incorporates ideas from deep learning based models directly into the state space model framework.
\end{enumerate}

\section{Stanza: a State Space Model with Nonlinear Activations and Dynamically Weighted Latent (Z) States}\label{sec:model}

Stanza is designed for competitive performance on both inference and forecasting tasks, specifically for settings with limited data and/or where parameter uncertainty estimates and interpretable models are critical.  Building off ideas from traditional state space models, particularly Time-Varying Autoregressive (TVAR) models \citep{West-Prado:1999}, Stanza combines ideas from deep learning approaches into the state space model framework for an intermediate approach that balances advantages of probabilistic inference and competitive forecasting accuracy. 



\subsection{Model}

Let $\{X_t\}$ be a univariate or multivariate time series of length $T$ centered about 0, and $\{Z_t\}$ a sequence of real-valued latent states.  The Stanza model is specified in  Equation~\ref{Stanza}:
\begin{equation}\label{Stanza}
\begin{split}
Z_t & \sim \mathcal{N}\left(f\left(\bm{W}_t^T\bm{Z}_{t-r:t-1}\right), \; Q_t\right), \\
X_t & \sim \mathcal{N}\left(HZ_t, \; R\right), \\
\bm{W}_t & \sim \mathcal{N}\left(\bm{W}_{t-1}, \; \bm\Lambda_t\right), \\
\end{split}
\end{equation}

where $\bm{Z}_{t-r:t-1} = [Z_{t-r}, \ldots, Z_{t-1}]^T$.  The dynamic weights, $\bm W_t$, allow for non-stationarity in the modeling of the time series and can directly capture structure in the observed time series.\footnote{Note: for clarity of notation, when $t < r$, there are 0s in the weight vector $\bm W_t$ for the elements corresponding  to times $t < 1$.  For example, if $r = 3$ and $t = 2$, then $\bm W_2 = [0, 0, W_1]^T$.}  The hyper-parameter $r$ determines the order of the lag dependence in the state space and can be specified directly with knowledge about the application.  For example, for daily time series data, if there is known weekly seasonality, then $r$ can be set to 7 and the dynamic weights, $W_{t-7}$ are expected to play an important role in the modeling of the time series.  The lag dependence can also be specified via cross-validation, allowing for structure to be learned from the time series. The lagged dependence enables series with longer term structure to be modeled as compared to the standard DLM or Kalman Filter, which utilize only a first-order Markov assumption in the latent space.  The function $f$ is the hyperbolic tangent function, which helps prevent the values of $Z_t$ from increasing without bound and also aids in constraining the model for identifiability.  This specification extends the basic  TVAR framework of \citep{West-Prado:1999} with ideas from attention-based models, such as \citep{NIPS2019_9311}, where the complexity of the model is captured in the state space, rather than the observed space, via dynamic weights.  In the case of Stanza, these dynamic weights, $\bm W_t$ can be directly interpreted in terms of structure in the observed time series.

Similar to deep models, Stanza specifies the model complexity in the latent space, rather than  the emission distribution, which is kept simple.  This setup allows for flexibility to model either univariate or multivariate series   $\{X_t\}$ within the same framework due to the simplicity of the emission distribution.  This is in contrast to TVAR models, which would require matrix-normal distributions to model multivariate sequences.  Additionally, in contrast to LSTMs, Stanza allows for probabilistic, interpretable inference, while still allowing for long-term dependencies, specified by $r$.  An external latent sequence can optionally be added to the state space to serve as an external input signal, as often used in traditional Kalman Filters \citep{Simon:2006} or to be used as a multi-scale latent factor in a decouple-recouple setting \citep{BerryWest2018DCMM, Ferreira2006, West-Harrison}.  

\subsection{Inference}

Filtering and smoothing of the latent space are critical to inference in DLMs and Kalman filters, and inference in Stanza proceeds similarly. The Extended Kalman Filter \citep{Ljung:1979} and Unscented Kalman Filter \citep{Wan-Merwe:2000} are extensions to the forward filtering algorithms in the standard Kalman Filter for estimation in the nonlinear setting.  We utilize both approaches in the filtering of parameters for Stanza. The Unscented Kalman Filter  uses a deterministic sampling approach to approximate the mean and variance of the state variables and can capture the mean and variance of the filtered latent state variable out to the third order \citep{Wan-Merwe:2000}.  The Extended Kalman Filter uses a first order Taylor approximation of the nonlinearity to update the traditional Kalman Filter \citep{Ljung:1979}.  Stanza is motivated as a competitive approach to existing deep and state space models and we thus want inference to be relatively fast and efficient.  Thus, we avoid the use of particle filters for inference due to their computational cost.  A comparison of the Unscented and Extended Kalman Filters is discussed in  \citep{Gustafsson:2012}.  Smoothing results in posterior distributions $p(Z_t | X_{1:T})$ for each time $t$. 

The inference procedure for Stanza is outlined in Algorithm~\ref{inf:Stanza} and mainly consists of nonlinear filtering and smoothing of the latent variables, followed by EM algorithm updates for the remaining parameters.   First, we perform Unscented Filtering \citep{Wan-Merwe:2000, Simon:2006} and Smoothing \citep{Sarkka:2008, Simon:2006} for the latent sequence $\{Z_t\}$.  Specifically, we specify Sigma Points, a minimal set of sample points, for all lagged $Z_t$ values for use in the Unscented Filter and Smoother to increase the fidelity of the approximation. We approximate the covariance matrix of these lagged values with a diagonal matrix of the filtered variances at previous time points.  Then, we perform Extended Kalman Filtering \citep{Ljung:1979, Simon:2006} and regular RTS Smoothing \citep{Rauch:1965} for the latent weight vectors $\bm W_t$.  There are no nonlinearities in the state equation for $\bm W_t$, so the Unscented Kalman Filter results in degenerate cross-covariance matrices, as the Sigma Points used in inference expect nonlinearities in the state space and observation space.  Thus, we use the Extended Kalman Filter combined with the TVAR inference algorithm from \citep{West-Prado:1999}, with corresponding first derivative corrections due to the nonlinearity.  

We initialize these latent variables via a ``warm start'' approach, which ignores the nonlinearity in the state space.  As filtering and smoothing depend on both $Z_t$ and $\bm W_t$, it is important to start with reasonable values of the latent variables for fast and accurate inference in the Stanza model.  We first initialize the $Z_t$ sequence by the traditional, linear Kalman Filter and RTS Smoother with randomly sampled, time-invariant parameters.  Then, we sample a latent trajectory from the smoothed estimates to initialize the $\bm W_t$ sequence via the TVAR inference algorithm \citep{West-Prado:1999}, which is similar and fully linear.  This warm start approach gives reasonable trajectories of the state variables $\{Z_t\}$ and $\{\bm W_t\}$ to start the Stanza inference procedure.

As it can be difficult to learn or to fully specify time-varying variances like $\{Q_t\}$ and $\{\bm{\Lambda_t}\}$, we use discount factors \citep{West-Harrison} to aid in the specification of these latent variances.  For a discount factor $\delta$, between 0 and 1 (but for most practical models, $\delta\in[0.9, 1)$), we specify $Q_t = \tfrac{1-\delta}{\delta}V_t$, where $V_t$ is the variance of $p(Z_t | X_{1:t})$ from forward filtering \citep{West-Harrison}.  The discount factor thus inflates the variance $Q_t$, relative to the filtered uncertainty $V_t$ and also represents a decay of information over time.  A similar discount factor setup can be specified for $\{\mathbf{\Lambda_t}\}$.  Lower discount factors allow for more volatility and adaptability of the filtered parameters (and hence forecast distributions), while higher discount factors result in much less inflation of the variance and less increase in the uncertainty from one time step to the next.  Discount factors are thus directly interpretable in terms of the increase in the uncertainty and adaptability of the parameters for filtering from one time step to the next, and can thus either be specified by domain knowledge or by cross-validation.  Discount factors are very general and can be extended to many settings; for a full discussion, see \citep{West-Harrison}.  We specify discount factors $\delta_Z$ and $\delta_W$ for   $\{Q_t\}$ and $\{\mathbf{\Lambda_t}\}$, respectively.

Finally, we learn the parameters $H$ and $R$ via the EM algorithm, following \citep{Khan-Dutt:2007}, where these updates have closed forms and use the mean and variance of the smoothed $\{Z_t\}$ from the Unscented Kalman Smoother.  Alternatively, priors could be specified for these parameters and $H$ and $R$ could be learned via Gibbs Sampling.  The simplicity of the emission distribution in Equation~\ref{Stanza} allows for simple updates of $H$ and $R$ and also enables equally simple inference for these parameters whether $X_t$ is a univariate or multivariate sequence. The convergence criteria of  Algorithm~\ref{inf:Stanza} can either be on the log-likelihood of the model or on the change in one of the EM parameters, $H$ or $R$.  Inference in Stanza is relatively fast and efficient; a full timing comparison with related approaches is discussed in Section~\ref{sec:results} and full inference details are given in the Supplementary Materials. 

There are three hyper-parameters to specify for Stanza: $r$, the lagged dependence in the latent space, $\delta_Z$, the discount factor for $\{Q_t\}$ and $\delta_W$, the discount factor for $\{\bm{\Lambda_t}\}$.  All of these parameters are directly interpretable and can either be specified directly via domain knowledge or selected using cross validation. 

\begin{algorithm}[H]\label{inf:Stanza}
\SetAlgoLined
\KwResult{Posterior distributions for $Z_{1:T}$ and $\bm W_{1:T}$, estimates $H$, $R$, $Q_{1:T}$, $\bm\Lambda_{1:T}$}
 Specify $r$, $\delta_Z$ and $\delta_W$\;
 Warm Start: Filter and smooth $Z_{1:T}$ via the Kalman Filter and RTS Smoother; filter and smooth $\bm W_{1:T}$ via the TVAR inference algorithm\;
 \While{While not converged}{
  Filter $Z_{1:T}$ via the Unscented Kalman Filter\;
  Smooth $Z_{1:T}$ via the Unscented Kalman Smoother\;
  Filter $\bm W_{1:T}$ via the Extended Kalman Filter; update $Q_t$ and $\bm\Lambda_t$ using discount factors\;
  Smooth $\bm W_{1:T}$ via the RTS Smoother\;
  Update $H$ and $R$ via the EM Algorithm. \
 }
 \caption{Stanza Inference}
\end{algorithm}


\subsection{Forecasting}

As in the basic DLM setting, forecasting in Stanza proceeds very naturally via simulation.  At a high-level, forecasting via simulation proceeds similarly to the regular filtering procedure in Algorithm~\ref{inf:Stanza}, but where values $Z_{t+1}$ and $X_{t+1}$ are sampled from the one step-ahead forecast distribution and used as the ``observed'' values in propagating the state variables forwards in time.  Specifically, for one-step ahead forecasting at time $t$, we first propagate the estimate for $\bm W_t$ through the state update equation via the Extended Kalman Filter, where the value for $Z_t$ at time t is sampled from the prior distribution.  Then, after completing the update from time $t-1$ to time $t$ for $\bm W_t$, we can sample a value of $\bm W_t$ from the Extended Kalman Filter update, that is used to update the estimate for $Z_t$ via the state update equation and the Unscented Kalman Filter.  After propagating both of the state space sequences forwards in time, a value of $X_t$ can easily be sampled from the measurement update equations in the Unscented Kalman Filter.  This facilitates probabilistic forecasting of $X_t$ and can be repeated to forecast $k$ steps ahead in time  
\citep{West-Harrison}.  It is also possible to propagate only the forecast mean, for example, rather than the full forecast distribution, for computational efficiency.

\section{Related Work}\label{sec:related_work}

Long Short-Term Memory (LSTM) models \citep{LSTM:1997} are widely used deep learning models for sequences or time series and address several of the training issues associated with recurrent neural networks \citep{Deep-Learning:2016}, and have been applied in many different applications.  However, LSTMs can still struggle to capture long-term structure in sequences and time series \citep{alex2019statistical} and more recently, attention-based approaches \citep{NIPS2017_7181, NIPS2015_5653} have achieved state-of-the-art results for sequence modeling.  While these models are capable of capturing long-term structure in time series, they require large amounts of data to train and are not necessarily suited to settings where probabilistic, interpretable inference is important or for settings in the small to medium data regime.  Autoregressive models such as \citep{sen2019think} are additionally well suited to large amounts of data, but not necessarily for settings where interpretable, probabilistic inference is required. 

Dynamic Linear Models (DLMs) \citep{West-Harrison}, which include Kalman Filters \citep{Kalman:1960}, are linear models that apply probabilistic filtering and smoothing.  Extensions include Dynamic Generalized Linear Models, which model $\{X_t\}$ from exponential family distributions \citep{West-Harrison, BerryWest2018DCMM}, and to include nonlinearities in the state space and observation equations \citep{Simon:2006, Wan-Merwe:2000, 6789612}.  Stanza extends ideas from TVAR models, in particular \citep{West-Prado:1999}.  DLMs and various extensions are discussed in detail in \citep{Roweis:1999} and in a Bayesian time series setting in \citep{West-Harrison}.  

Finally, there are several related approaches that try to address the range between full deep learning approaches and linear state space models, either by including probabilistic inference in deep learning approaches or by extending state space models to more complex settings.  Deep learning focused models include Deep Kalman Filters \citep{krishnan2015deep}, which use neural networks for the emission and transition distributions in the DLM specification, probabilistic recurrent state space models \citep{doerr2018probabilistic}, which combine RNN training with Gaussian processes and Gaussian process dynamic models \citep{Wang:2005} for motion capture data.  Additional prior work for probabilistic time series focuses on different aspects of improving inference in complex models  \citep{archer2015black, Karl:2017, Krishnan:2017}.  Finally, there is recent work focusing specifically on improving forecasting using deep models to extend state space models, such as \citep{NIPS2018_8004, NIPS2019_8907}.  

Related approaches that extend state space models directly to nonlinear or more complex settings include \citep{Raiko:2009}, which applies the variational inference approach proposed in \citep{6789612} to learn nonlinear state space models for model predictive control, rather than inference or forecasting, and \citep{NIPS2019_9311}, which uses a discrete state space and an attention mechanism learned via a sequence-to-sequence model to model disease progression.

Stanza, however, is distinct from these previously proposed approaches.  Unlike \citep{Raiko:2009, NIPS2018_8004}, Stanza is motivated by a desire to perform competitively for both inference \textit{and} forecasting settings.   Thus, Stanza is designed to be a \textit{versatile} model that can perform well under a variety of different circumstances.  Additionally, rather than applying state space ideas to deep learning approaches \citep{krishnan2015deep, doerr2018probabilistic, NIPS2018_8004, archer2015black}, Stanza incorporates ideas from deep learning into a state space framework.  Similar to \citep{NIPS2019_9311}, Stanza creates dynamic dependencies in the latent space using dynamic weighting, rather than an attention mechanism with a discrete state space \citep{NIPS2019_9311}, to do so.  Additionally, Stanza specifies complexity in the latent space and adds nonlinearities to improve performance.

\section{Experiments}\label{sec:results}
Stanza is designed for competitive performance in inference and forecasting settings.  We demonstrate Stanza's inference utility on two simulation experiments, both of which consider settings where Stanza is able to capture relevant features in the data, while traditional DLMs are not sufficiently rich to do so.  We also demonstrate Stanza's competitive forecasting accuracy on three real-world datasets.  

\subsection{Simulations}

\begin{figure}
\centering
\begin{subfigure}{.5\textwidth}
  \centering
  \includegraphics[width=\linewidth]{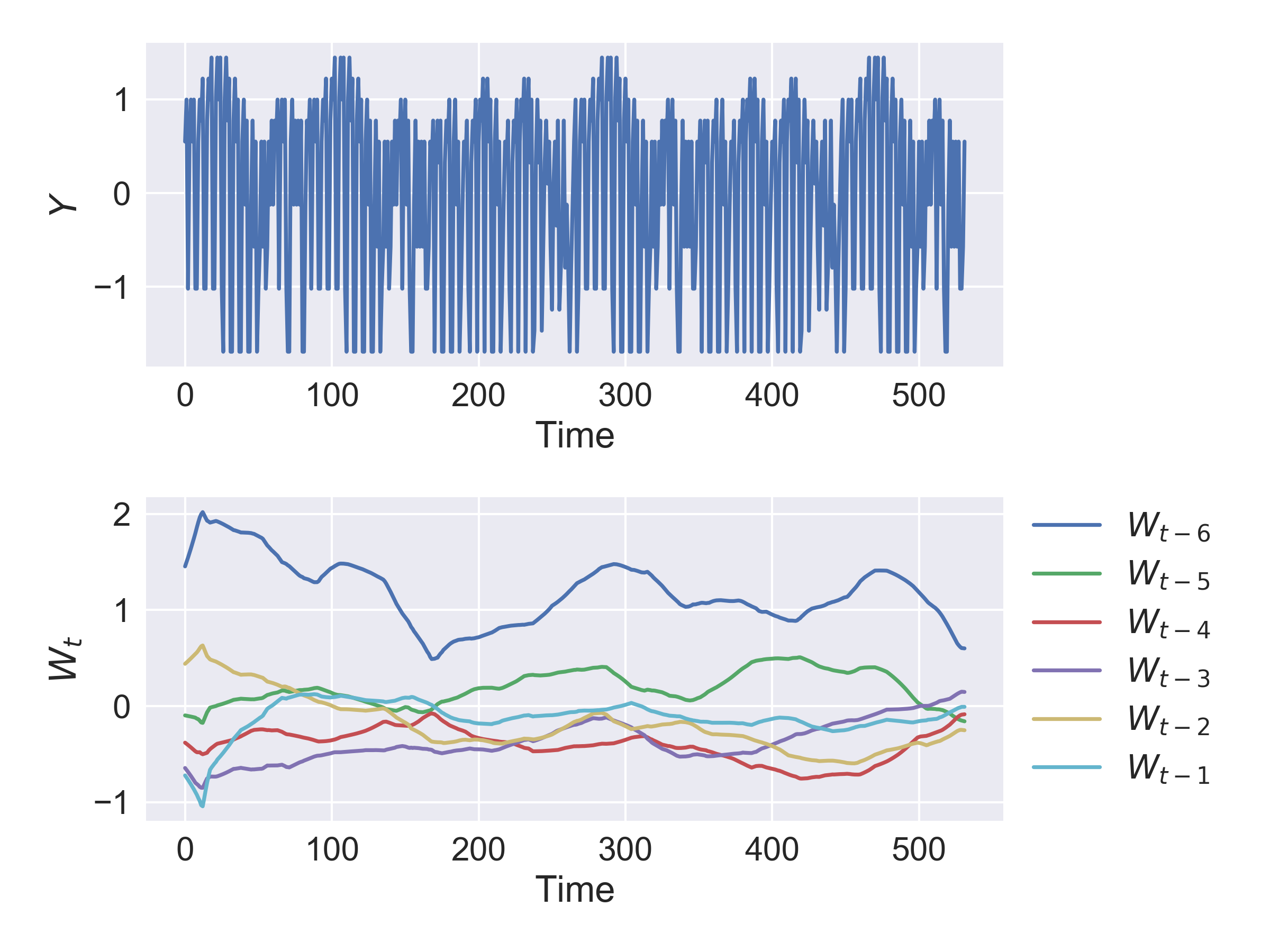}
  \caption{$Y$ and $\bm W_t$.}
  \label{fig:Y}
\end{subfigure}%
\begin{subfigure}{.5\textwidth}
  \centering
  \includegraphics[width=\linewidth]{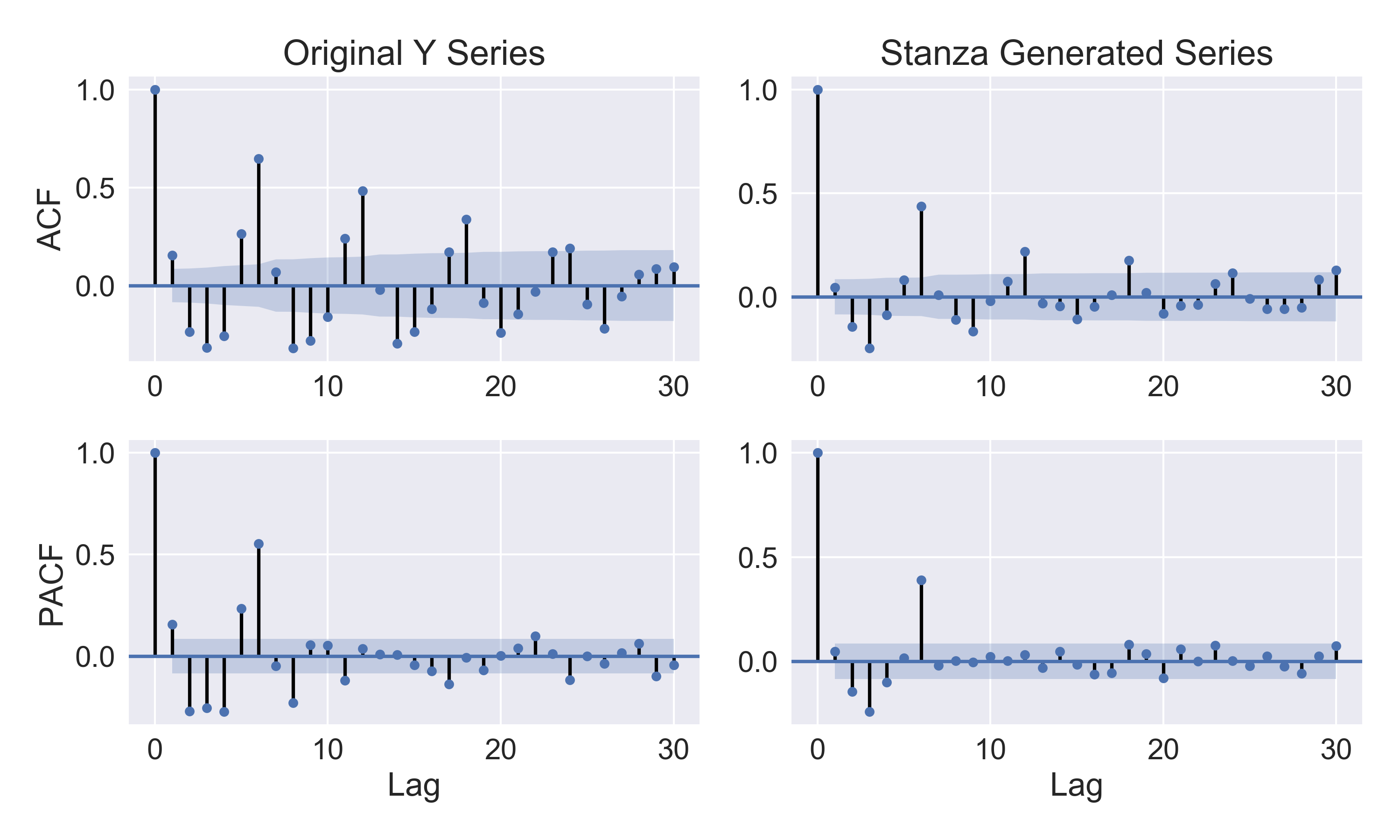}
  \caption{ACF and PACF plots.}
  \label{fig:ACF}
\end{subfigure}
\caption{For a highly structured, periodic time series $Y$ (a, upper) with strong seasonality, as evidenced in the ACF/PACF plots (b, left), Stanza is able to recover this structure in the $\bm W_t$ parameters.  i.e. $W_{t-6}$ is important and changes when the pattern in $Y$ changes (a, lower).  Additionally, when generating from the learned Stanza model, the generated series is able to exhibit the same structure as the original series, in terms of the ACF and PACF (b, right).}
\label{fig:Ode}
\end{figure}

We first demonstrate the ability of Stanza to recover structure in time series.  We consider a highly structured series (Figure~\ref{fig:Y} top with autocorrelation function (ACF) and partial autocorrelation function (PACF) plots Figure~\ref{fig:ACF}).  There is clear seasonality out to lag 6, so we can specify $r=6$ directly in the Stanza model.  The resulting posterior mean weight matrix, $\bm W_t$ is shown in the bottom of Figure~\ref{fig:Y}.  The weight corresponding to times $t-6$ is the largest across time, reflecting the importance of the seasonality evident in the original series.  Additionally, the structure of the time series changes slightly between times 150-250 and 300-400, and this change is evident in decreased values of the $W_{t-6}$ weights at these same times (Figure~\ref{fig:Y} bottom).  Additionally, we reproduce this structure when we simulate from
from the learned Stanza model, see the ACF/PACF plots in Figure~\ref{fig:ACF} right.  Thus, for a highly structured time series, Stanza is able to recover key features of the structure and dependence of the series over time in the dynamic weight matrix $\bm W_t$, which can be directly interpreted in the context of the application.

\begin{figure}
\centering
\begin{subfigure}{.63\textwidth}
  \centering
  \includegraphics[width=\linewidth]{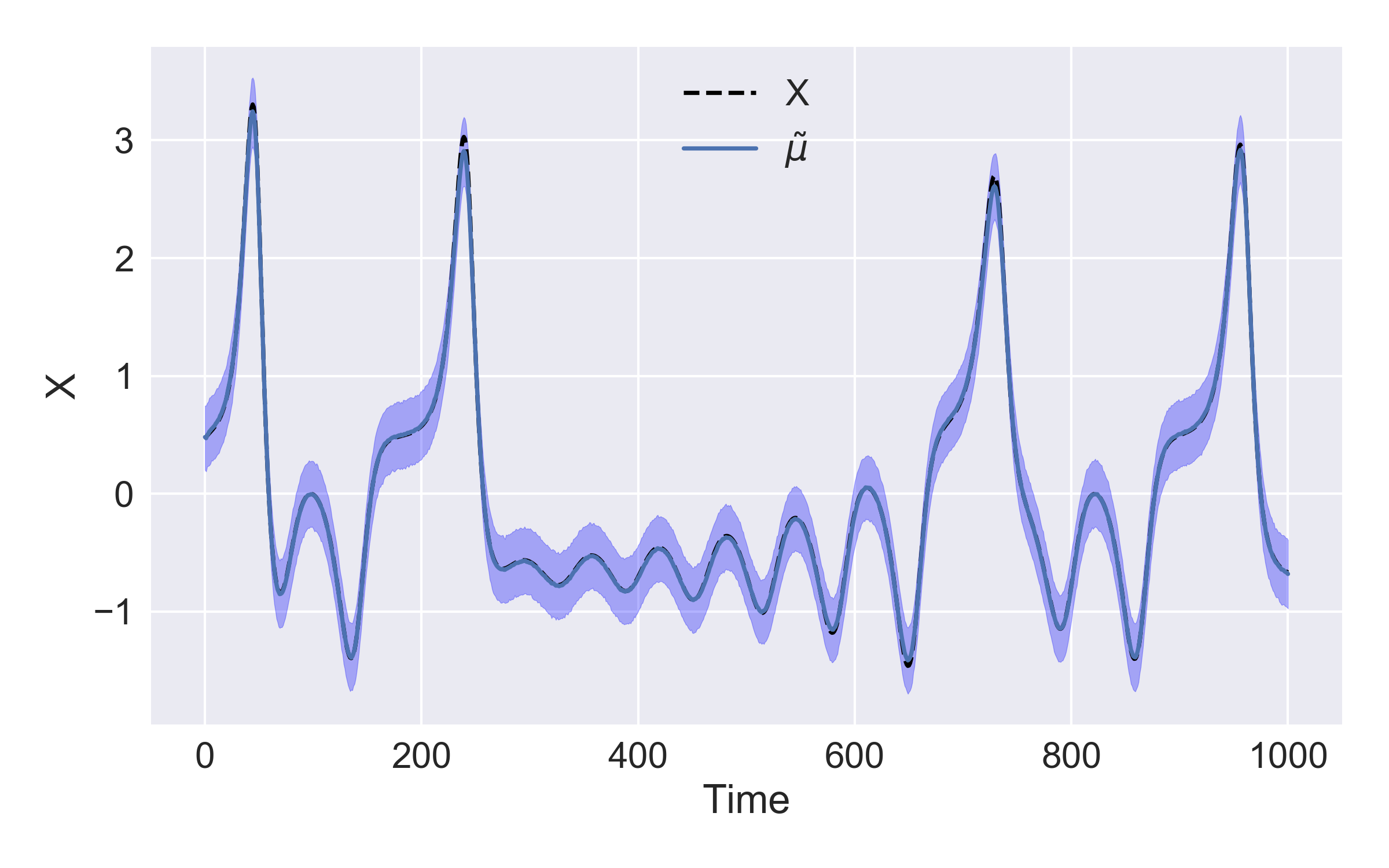}
  \caption{Posterior predictive mean.}
  \label{fig:Lorenz-post}
\end{subfigure}%
\begin{subfigure}{.37\textwidth}
  \centering
  \includegraphics[width=\linewidth]{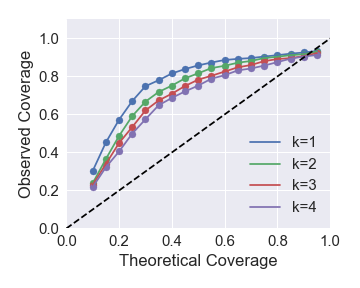}
  \caption{Coverage plots.}
  \label{fig:Lorenz-coverage}
\end{subfigure}
\caption{(a) Posterior predictive distribution for a Stanza(5) model fit to the $x$ trajectory for a Lorenz attractor.  The posterior predictive mean, $\tilde{\mu}$, agrees very well with the observed trajectory. (b) Forecast distribution coverage for $k = 1, \ldots, 4$ step-ahead forecasts for the Stanza(5) model fit to the Lorenz attractor trajectory. }
\label{fig:Lorenz}
\end{figure}

To demonstrate Stanza's ability to model complex, nonlinear time series, we generate data from a Lorenz attractor \citep{Lorenz:1963}, following \citep{NIPS2019_9532}.  A Lorenz attractor is defined by a system of three differential equations and is an example of a chaotic system.  We    
simulate 1000 time points from a Lorenz attractor with a time delta of 0.01, and consider the trajectory of the first variable, $x$.  We then fit a Stanza(5) (i.e. $r=5$) model with high discount factors of 0.99 for $\delta_W$ and $\delta_Z$, indicating little volatility.  The posterior predictive distribution for the learned Stanza model is shown in Figure~\ref{fig:Lorenz-post}. There is excellent correspondence between the posterior predictive mean and the observed Lorenz attractor trajectory, as well as good coverage for multi-step ahead forecasting (Figure~\ref{fig:Lorenz-coverage}).  As Stanza performs probabilistic forecasting, we can calculate the coverage of the forecast distribution as the percentage of observed values that fall within specific percentiles of the forecast distribution.  That is, we expect approximately 95\% of the observed values to fall within the 95\% credible intervals for the forecast distribution.   

\subsection{Forecasting}\label{sec:forecasting}

We additionally demonstrate Stanza's competitive forecasting performance on three real-world datasets.  Following \citep{NIPS2019_8907, Lai:2017, sen2019think}, we compare Stanza's forecasting performance on the following publicly available datasets: Exchange Rate \citep{Lai:2017},  the daily exchange rate for 8 different currencies from 1990-2011, and Electricity\footnote{\url{https://archive.ics.uci.edu/ml/datasets/ElectricityLoadDiagrams20112014}}, hourly electricity consumption for 370 customers (subset to the first 50 customers and the last 10000 time points).    We additionally analyze the Weather dataset from \citep{Chollet:2017,Tutorials:2020}, which contains 13 different weather measurements.  We subset the Weather dataset to analyze the data from 10/01/2016 to 12/31/2016. For each forecasting experiment considered below, the last 1000 time points of each series are treated as test data and the preceding points as the training data.  All series are individually normalized using the mean and standard deviation of the training set.  

\begin{figure}
\centering
\begin{subfigure}{.4\textwidth}
  \centering
  \includegraphics[width=\linewidth]{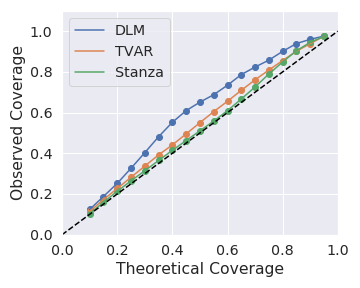}
  \caption{Exchange Rate.}
  \label{fig:ex}
\end{subfigure}%
\begin{subfigure}{.4\textwidth}
  \centering
  \includegraphics[width=\linewidth]{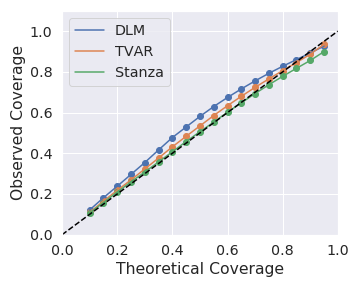}
  \caption{Electricity.}
  \label{fig:elec}
\end{subfigure}
\caption{Coverage for $k=10$ step-ahead forecasting on the (a) Exchange Rate and (b) Electricity datasets. Stanza achieves excellent coverage on both datasets. }
\label{cov}
\end{figure}

We fit several models independently to each series in each dataset and compare forecasting accuracy with the root-mean squared error (RMSE) for $k=1$, $k=5$ and $k=10$ step-ahead forecasting, where $k$ is the forecast horizon.  We compare Stanza to two baseline models, a normal DLM with time-invariant parameters \citep{West-Harrison} and a TVAR model \citep{West-Prado:1999, Prado-West}, as well as two LSTMs, one with 2 hidden layers (LSTM(2)) and one with 5 hidden layers (LSTM(5)).  For a direct comparison across series, we use cross-validation to tune the three hyper-parameters for Stanza (using one-step ahead forecasting RMSE as a metric) on each dataset and use the same hyper-parameters for all series for both the Stanza and TVAR models for consistency.  As Exchange Rate is a daily dataset, the optimal value of $r=7$ for Stanza represents weekly seasonality, and Stanza is able to capture relevant structure in the data.  To train the LSTMs, we use a rolling window approach, where the previous 50 data points are used to predict the next 10.  The LSTMs are trained with the Adam optimizer on MSE loss.  Additionally, as all models except for the LSTMs make probabilistic forecasts, we compare the 95\% forecast coverage of the baseline models to Stanza.

The forecasting results for the individual series are given in Table~\ref{forecast:individ}.  For all three datasets and across all three forecast horizons, Stanza achieves a good balance of high forecast accuracy and good empirical coverage (Figure~\ref{cov}).  As compared to competing approaches, Stanza's advantages are the ability to model multivariate time series (unlike TVAR), to produce probabilistic forecasts and thus coverage estimates (unlike deep LSTMs) and to model non-stationary time series (unlike DLMs).  The forecasting accuracy of Stanza does not decay over longer forecasting horizons, and is competitive with the performance of the deep LSTMs across all forecast horizons.  Additionally, LSTMs do not provide probabilistic forecast distributions and thus cannot be used to calculate coverage of the observed time series, while Stanza achieves excellent coverage   (Figure~\ref{cov}), even compared to the DLM and TVAR models.

\begin{table}
  \caption{Forecast performance of Stanza as compared to several baseline models.  Average RMSE and empirical 95\% coverage, where the results are averaged across series. $k$ denotes the forecast horizon. The numbers in parentheses refer to the order of the model; Stanza(7) is a Stanza model where $r=7$ and LSTM(2) is an LSTM with two layers.}
  \label{forecast:individ}
  \centering
  \scalebox{0.88}{
  \begin{tabular}{llrrrrrr}
    \toprule
 & &   \multicolumn{3}{c}{RMSE}    &  \multicolumn{3}{c}{95\% Coverage}               \\
\cmidrule(lr){3-5} \cmidrule(lr){6-8} 
       Dataset     & Model     & $k = 1$ & $k = 5$ & $k=10$ & $k = 1$ & $k = 5$ & $k=10$ \\
    \midrule
   \multirow{3}{*}[-2pt]{Exchange Rate} & DLM  & $0.91\pm 0.34$ &	$0.90\pm0.34$&	$0.90\pm0.34$& $98$	&$98$&	$98$    \\
         & TVAR(7) &  $0.05\pm 0.02$	& $0.14\pm 0.17$	& $11.19\pm 29.32$ &  $94$	&$96$	&$98$ \\
         & LSTM(2)       &  $0.47\pm 0.22$	&$0.48\pm 0.21$&	$0.49\pm0.20$ & \textemdash & \textemdash & \textemdash \\
         & LSTM(5)       &  $0.49\pm0.23$&	$0.51\pm0.24$&$0.54\pm 0.25$ & \textemdash & \textemdash & \textemdash \\
         & \textbf{Stanza(7)}       &   $0.39\pm0.22$&	$0.47\pm0.28$	&$0.58\pm 0.37$ & $100$&$100$&	$97$ \\
    \midrule
    \multirow{3}{*}[-2pt]{Electricity}  & DLM  &   $1.16\pm 0.24$&	$1.16\pm 0.24$&	$1.17\pm0.24$ & $93$	&$93$&	$93$  \\
         & TVAR(5) &   $0.45\pm0.12$	& $1.06\pm0.25$&	$1.15\pm0.26$ &  $93$ &	$92$	&$94$ \\
         & LSTM(2)       &  $0.82\pm0.24$ &	$0.73\pm0.20$ &	$0.93\pm0.24$ & \textemdash & \textemdash & \textemdash \\
         & LSTM(5)       &  $0.55\pm 0.18$&	$0.66\pm 0.21$	&$0.70\pm0.21$ & \textemdash & \textemdash & \textemdash \\
         & \textbf{Stanza(5)}       &  $0.89\pm0.25$	& $1.15\pm 0.27$&	$1.17\pm0.27$& $93$	&$90$&	$90$ \\
         \midrule
    \multirow{3}{*}[-2pt]{Weather}  & DLM  &   $1.01\pm0.22 $&	$1.00\pm0.22 $&	$1.00\pm0.22$ & 97	&97&97	 \\
         & TVAR(6) &   $0.11\pm0.15$	& $0.19\pm0.17$&	$0.25\pm0.18$ &  95 &	96	& 96\\
         & LSTM(2)       &  $0.52\pm0.17$ &	$0.57\pm0.15$ &	$0.61\pm0.15$ & \textemdash & \textemdash & \textemdash \\
         & LSTM(5)       &  $0.65\pm 0.37$&	$0.71\pm 0.34$	&$0.75\pm0.33$ & \textemdash & \textemdash & \textemdash \\
         & \textbf{Stanza(6)}       &  $0.58\pm0.18$	& $0.72\pm 0.19$&	$0.82\pm0.21$& 99	&98&	97 \\
       \bottomrule
  \end{tabular}
  }
\end{table}



The TVAR model achieves very good forecasting accuracy, across the datasets.  However, the TVAR model is not designed for multivariate series, while Stanza is easily extended to multivariate $X_t$ sequences.  We fit Stanza to the full multivariate Exchange Rate and Weather datasets, due to their relatively low dimension.  These results are given in Table~\ref{forecast:multi}.  This extension to the full multivariate sequence further improves the forecasting ability of Stanza, especially compared to the TVAR models and the deep LSTMs in Table~\ref{forecast:individ}, on the Exchange Rate data.

Finally, we can look at the computational efficiency of the Stanza model. For the Exchange Rate data, the average time in seconds for training Stanza is 41 seconds, while for the LSTM(5) model is 222 seconds (averaged across series, run on a single core).  While forecasting via simulation for Stanza is much slower than for the LSTM, this computational time could be reduced via propagating the mean only or via parallelization.   


\begin{table}[h]
  \caption{Forecast performance of Stanza on multivariate sequences.}
  \label{forecast:multi}
  \centering
  \scalebox{0.88}{
  \begin{tabular}{llrrr}
    \toprule
 & &   \multicolumn{3}{c}{RMSE}                  \\
\cmidrule(lr){3-5} 
       Dataset     & Model     & $k = 1$ & $k = 5$ & $k=10$  \\
    \midrule
   Exchange Rate & \textbf{Stanza(7)}       &   0.09 &	0.09	& 0.09  \\
    \midrule
    Weather  & \textbf{Stanza(6)}       &  0.81	&  0.91&	0.92  \\
       \bottomrule
  \end{tabular}
  }
\end{table}

\section{Discussion}

Stanza is motivated by a general goal of integrating modern machine learning approaches with traditional stochastic models.  Specifically, we propose Stanza as a versatile, non-stationary state space model for competitive performance on inference and forecasting tasks.  Stanza is able to achieve competitive forecasting accuracy with deep LSTMs, while still maintaining probabilistic estimates of the state space  and interpretability of parameters.  Stanza is suitable for settings where parameter interpretability is important for modeling complex time series and is suitable for either univariate or multivariate modeling.   There is a long history of simple machine learning models, like random forests, outperforming deep models on certain tasks, and Stanza fits into this paradigm.  Future directions include extending these ideas further in the direction of deep learning models, for example by adding additional depth to the latent space in the proposed Stanza model or extensions for higher-dimensional time series, for example with ideas from \citep{sen2019think, NIPS2019_8907}.  Stanza is a step towards a powerful, general framework of incorporating deep learning ideas directly into traditional probabilistic models.


\bibliographystyle{abbrvnat}
\bibliography{refs}

\break\newpage
\appendix

\section{Inference}\label{sec:inference}

Let $\{X_t\}$ be a univariate or multivariate time series of length $T$ centered about 0, and $\{Z_t\}$ a sequence of real-valued latent states.  The Stanza model is specified in  Equation~\ref{Stanza2}:
\begin{equation}\label{Stanza2}
\begin{split}
Z_t & \sim \mathcal{N}\left(f\left(\bm{W}_t^T\bm{Z}_{t-r:t-1}\right), \; Q_t\right), \\
X_t & \sim \mathcal{N}\left(HZ_t, \; R\right), \\
\bm{W}_t & \sim \mathcal{N}\left(\bm{W}_{t-1}, \; \bm\Lambda_t\right), \\
\end{split}
\end{equation}

where $\bm{Z}_{t-r:t-1} = [Z_{t-r}, \ldots, Z_{t-1}]^T$.  Additionally, $Z_1 \sim\mathcal{N}\left(\mu_0, \; V_0\right)$ and $\bm W_1 \sim\mathcal{N}\left(\bm m_0, \; \bm C_0\right)$, where we specify $\mu_0 = 0, V_0 = 1, \bm m_0 = \bm 0, \bm C_0 = I_r$, where $I_r$ is the identity matrix of dimension $r$.  Additionally, in the TVAR model framework \citep{West-Prado:1999}, we specify $k_0 = 1, d_0 = 0.01$ for the volatility for $Z_t$; see \citep{West-Prado:1999, West-Harrison} for details.  

Filtering and smoothing of $\{Z_t\}$ proceeds via the Unscented Kalman Filter \citep{Wan-Merwe:2000} and Unscented Smoother \citep{Sarkka:2008, Simon:2006}.  Inference for $\{\bm W_t\}$ proceeds via the Extended Kalman Filtering \citep{Ljung:1979, Simon:2006} and regular RTS Smoothing \citep{Rauch:1965}.  The EM updates for the parameters $H$ and $R$ are given in Equation~\ref{uni} for univariate $X_t$ and Equation~\ref{multi} for multivariate $X_t$, where the expectations $\mathbb{E}_{Z|X}$  are obtained via the smoothing relationships for $\{Z_t\}$, as smoothing results in posterior distributions $p(Z_t | X_{1:T})$ for each time $t$.

\begin{equation}\label{uni}
\begin{split}
 \hat{H}_{EM} &= \dfrac{\sum_{t=1}^TX_t\mathbb{E}_{Z|X}(Z_t)}{\sum_{t=1}^T \mathbb{E}_{Z|X}(Z_t^2)} \\
\hat{R}_{EM} &= \dfrac{1}{T} \sum_{t=1}^T \left(X_t^2 - 2H_{EM} X_t\mathbb{E}_{Z | X}(Z_t) + H_{EM}^2\mathbb{E}_{Z|X}(Z_t^2)\right) \\
\end{split}
\end{equation}

\begin{equation}\label{multi}
\begin{split}
H_{EM} &= \dfrac{\sum_{t=1}^T X_t \mathbb{E}_{Z|X}(Z_t)}{\sum_{t=1}^T \mathbb{E}_{Z|X}(Z_t^2)} \\
R_{EM} &= \dfrac{1}{T}\sum_{t=1}^T \left[X_tX_t^T - 2H_{EM}X_t^T\mathbb{E}_{Z|X}(Z_t) + H_{EM}H_{EM}^T\mathbb{E}_{Z|X}(Z_t^2)\right]
\end{split}
\end{equation}





\section{Simulation Experiments}\label{sec:simulations}

Additional details about the Simulation Experiments in the main paper are discussed in this section.

\subsection{Kalman Filter Recovery}

\begin{figure}
  \centering
   \includegraphics[width=0.7\linewidth]{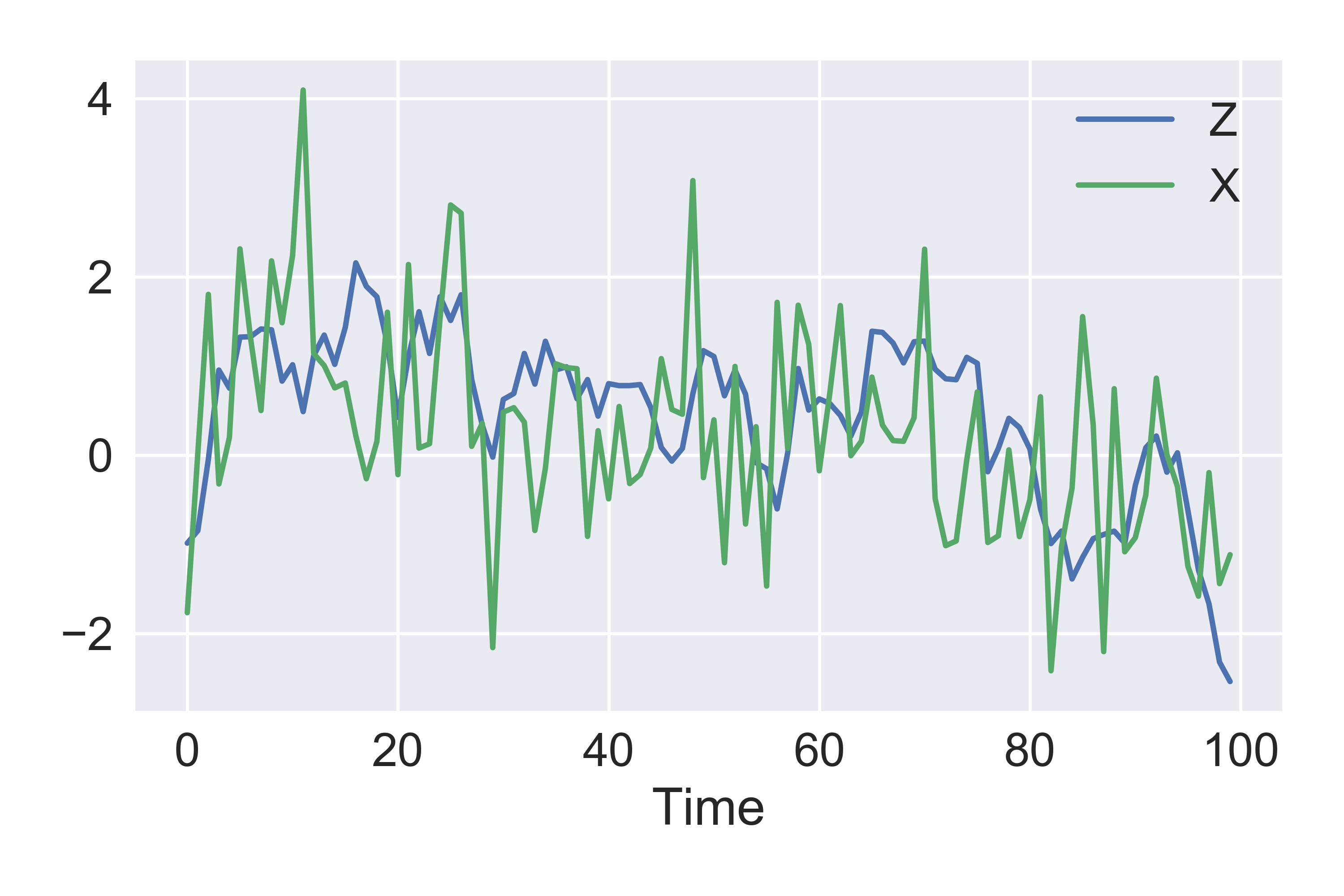}
  \caption{Latent, $\{Z_t\}$ and observed, $\{X_t\}$ sequences generated from a time-invariant Kalman Filter.  The parameters for this Kalman Filter are $F = 0.9$, $Q = 0.3$, $H = 0.5$ and $R = 1$.}
  \label{fig:Kalman}
\end{figure}

Stanza can also be fit to time series that follow dynamics that are simpler than those assumed in the Stanza model.  We generate data from a regular Kalman Filter with time-invariant parameters and then fit a Stanza(3) model with discount factors $\delta_Z = \delta_W = 0.99$.  The generated series $\{X_t\}$ and the generating latent sequence are shown in Figure~\ref{fig:Kalman}.  A time invariant Kalman Filter can be expressed as in Equation~\ref{Kalman} \citep{Kalman:1960, Simon:2006}:

\begin{equation}\label{Kalman}
\begin{split}
Z_t & \sim \mathcal{N}\left(FZ_{t-1}, \; Q\right), \\
X_t & \sim \mathcal{N}\left(HZ_t, \; R\right), \\
\end{split}
\end{equation}

\begin{figure}
  \centering
   \includegraphics[width=0.7\linewidth]{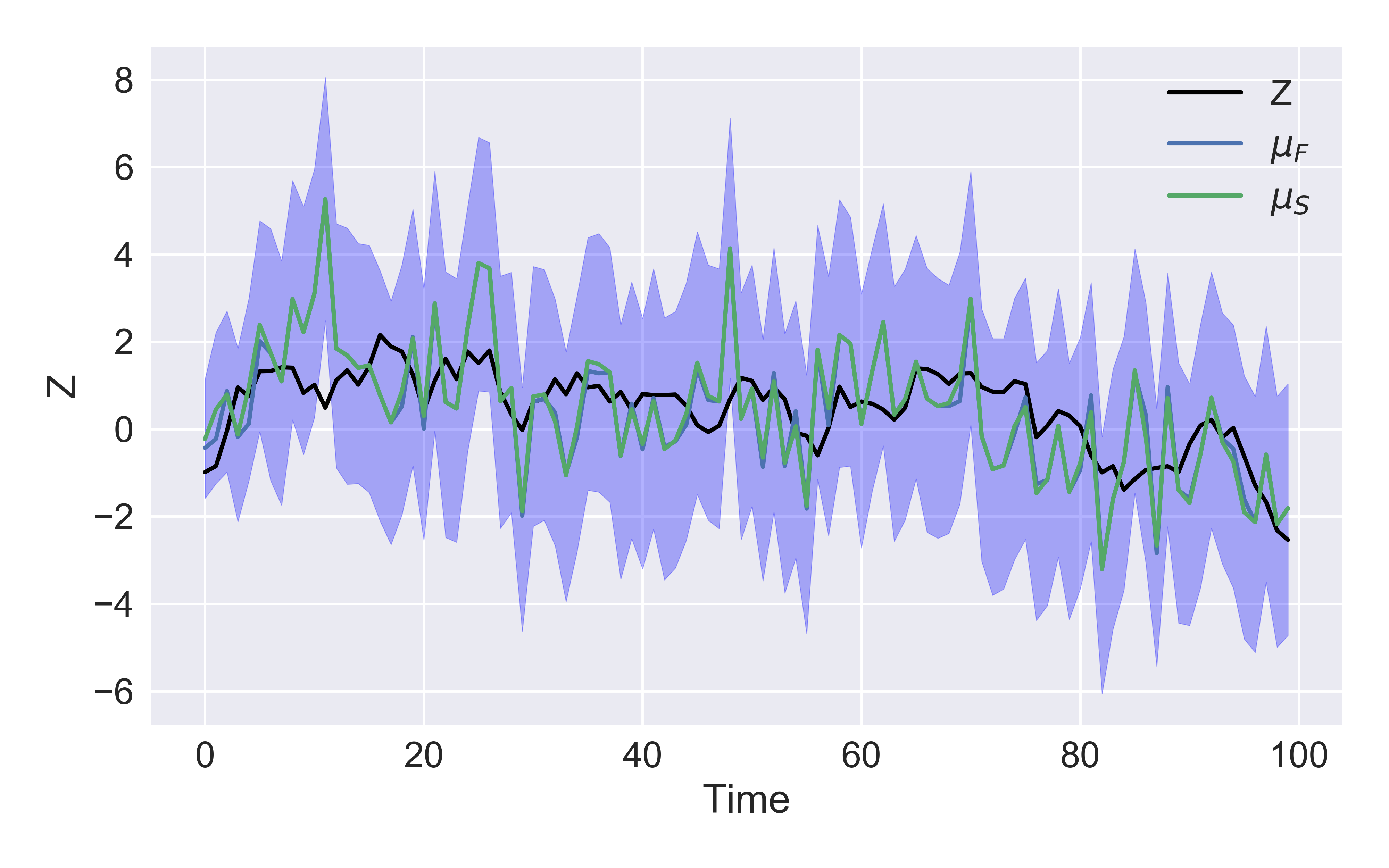}
  \caption{Filtered ($\mu_F$) and smoothed ($\mu_S$) means from Stanza for the Kalman Filter generated data.  The generating latent sequence is shown in black.  The shaded region shows the 95\% credible intervals using the smoothed mean and variance from the Unscented Kalman Filter.}
  \label{fig:Kalman-mu}
\end{figure}

\begin{figure}
  \centering
   \includegraphics[width=0.7\linewidth]{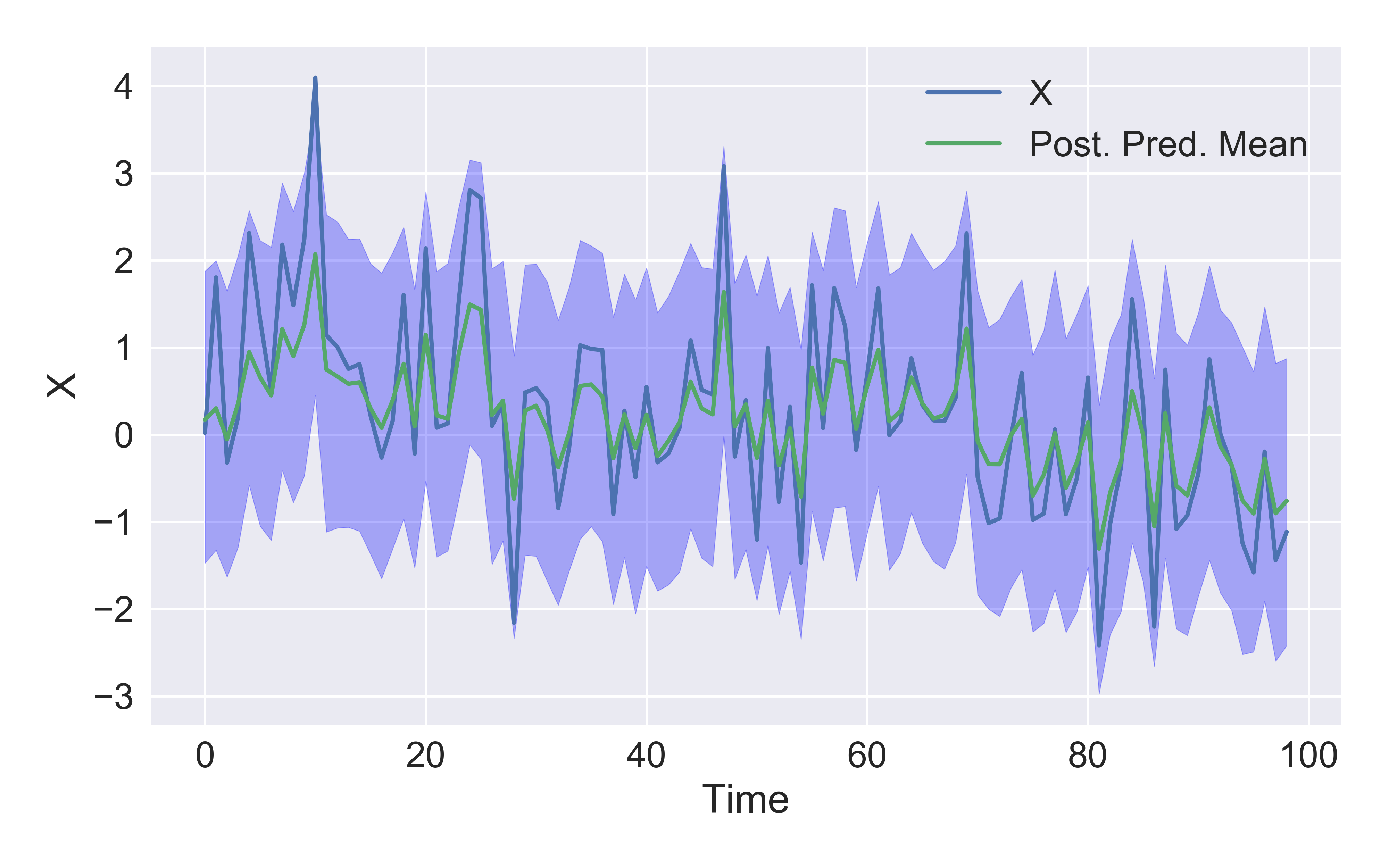}
  \caption{True $\{X_t\}$ sequence and posterior predictive mean from the Stanza(3) model on the Kalman Filter Data.  The shaded regions represent 95\% credible intervals. There is good correspondence and coverage of the Stanza model fit.}
  \label{fig:Kalman-postpred}
\end{figure}

The filtered, $\mu_F$, and smoothed, $\mu_S$, means from Stanza are shown in Figure~\ref{fig:Kalman-mu} and correspond well to the generating latent sequence, $\{Z_t\}$.  Finally, the posterior predictive mean for $\{X_t\}$ is shown in Figure~\ref{fig:Kalman-postpred} with credible intervals.  There is good correspondence between the observed, generated sequence $\{X_t\}$ and the Stanza model fit.  Thus, Stanza is able to model simpler dynamics then specified in the full model well.

\subsection{Structure Recovery}

\begin{figure}
  \centering
   \includegraphics[width=0.8\linewidth]{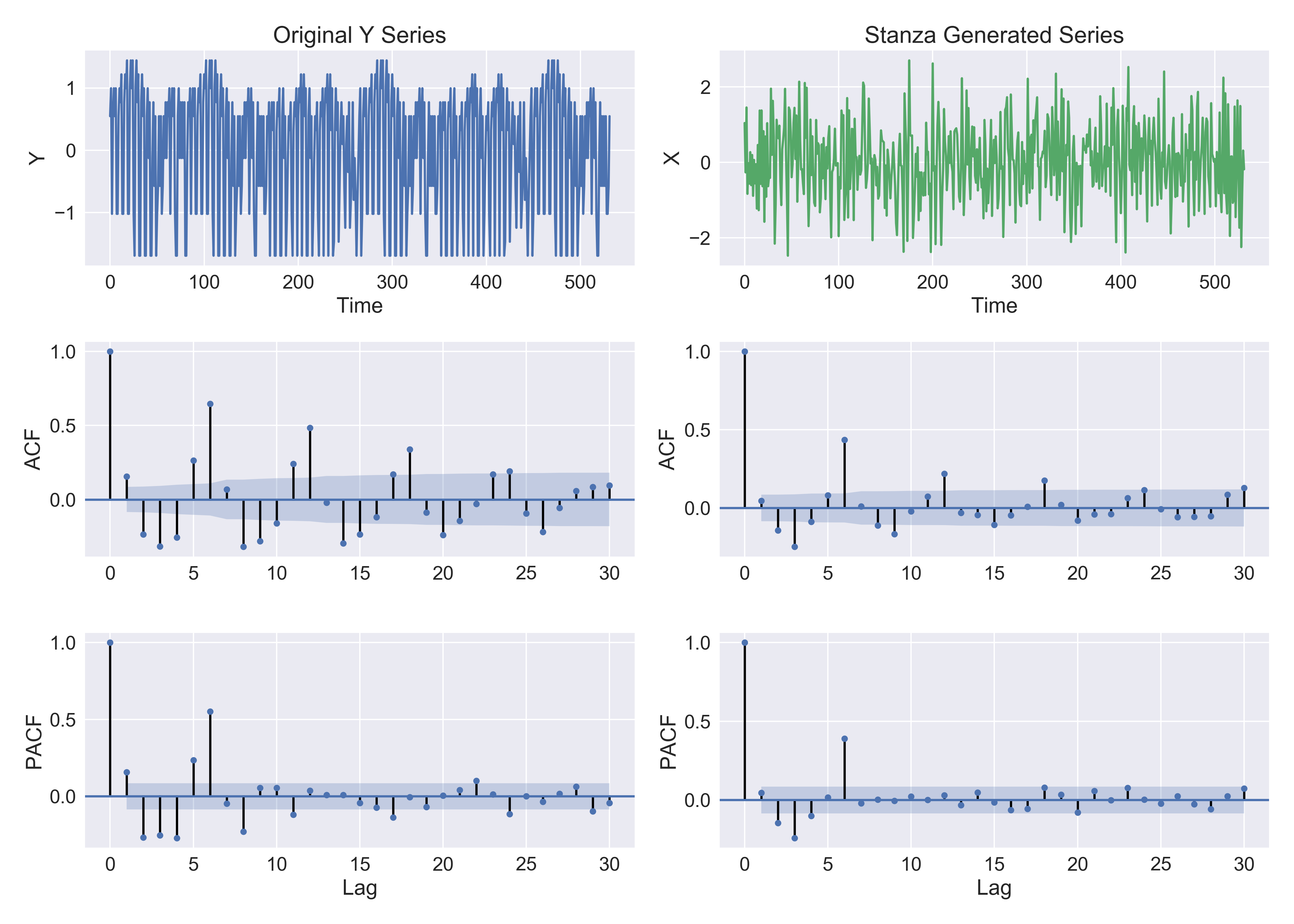}
  \caption{Left, original, highly periodic series plotted with ACF and PACF plots. There is clear seasonal structure, with period 6.  Right, time series generated by a Stanza(6) model on the time series on the left.  The structure in the ACF and PACF plots is recovered well by the Stanza(6) generated series.}
  \label{fig:Wt}
\end{figure}

A Stanza(6) model with $\delta_Z = \delta_W = 0.97$ is fit to highly periodic, structured data.  The learned model parameters are then used to generate new data, which corresponds well in terms of temporal structure to the original data (Figure~\ref{fig:Wt}).

\subsection{Lorenz Attractor Details}

To demonstrate Stanza's ability to model complex, nonlinear time series, we generate data from a Lorenz attractor \citep{Lorenz:1963}, following \citep{NIPS2019_9532}.  A Lorenz attractor is defined by a system of three differential equations and is an example of a chaotic system. The Lorenz equations used to generate the data for this simulation are given in Equation~\ref{Lorenz}.   We    
simulate 1000 time points from a Lorenz attractor with a time delta of 0.01, and consider the trajectory of the first variable, $x$.  We then fit a Stanza(5) model with high discount factors of 0.99 for $\delta_W$ and $\delta_Z$, indicating little volatility. 

\begin{equation}\label{Lorenz}
\dfrac{\partial x}{\partial t} = 10(y - x), \enspace \dfrac{\partial y}{\partial t} = x(28 - z) - y, \enspace \dfrac{\partial z}{\partial t} = xy - \dfrac{8}{3}z 
\end{equation}

\section{Forecasting Experiments}

Additional details about the forecasting experiments considered in the main paper are included here. 

\subsection{Datasets}

We demonstrate Stanza's competitive forecasting ability on three real world, publicly available datasets, following \citep{NIPS2019_8907, Lai:2017, sen2019think}.  Additional details about each dataset  are given in Table~\ref{tab:data}.   All subsetting is performed for computational efficiency only.  The Electricity\footnote{\url{https://archive.ics.uci.edu/ml/datasets/ElectricityLoadDiagrams20112014}} data is subset to the first 50 customers and the last 10000 time points.  The Weather   data is subset for the time range 10/01/2016 to 12/31/2016.  For all datasets and for each forecasting experiment considered below, the last 1000 time points of each series are treated as test data and the preceding points as the training data.  All series are individually normalized using the mean and standard deviation of the training set.  For the multivariate modeling, the Weather dataset is subset to 9 series.

\begin{table}
  \caption{Dataset details for the forecasting experiments.}
  \label{tab:data}
  \centering
  \scalebox{0.88}{
  \begin{tabular}{lrrr}
    \toprule
       Dataset    & Frequency & Dimension     & Number of Time Points   \\
       \midrule
   Exchange Rate \citep{Lai:2017} & Daily      &   8 &	7588 	 \\
   Electricity & Hourly & 50 & 10000  \\
    Weather \citep{Chollet:2017,Tutorials:2020}  & 10 mins       &  14	& 12804  \\
       \bottomrule
  \end{tabular}
  }
\end{table}

\subsection{Forecasting Experiment Hyper-Parameter Details}

For the univariate forecasting experiment, the hyper-parameters for the Stanza models are tuned on an aggregate series for the Electricity and Exchange Rate datasets, and then the same hyper-parameters are used for modeling all the individual series for the Stanza and TVAR models.  The hyper-parameters are selected based on which combination gives the best 1-step ahead forecasting RMSE, using a grid of values.  The hyper-parameters for the Electricity data are $r = 5, \delta_Z = \delta_W = 0.97$ and for the Exchange Rate data are $r = 7, \delta_Z = \delta_W = 0.97$. For the Weather dataset, as an aggregate series does not make sense (each individual series measures a different physical quantity on a different scale), we manually select $r = 6$ (corresponding to hourly structure in the data) and $\delta_Z = \delta_W = 0.97$, allowing for a relatively large amount of volatility in the measurements.

\subsection{Timing}

Full timing details for training and forecasting on the Exchange Rate data are given in Table~\ref{timing}.

\begin{table}[h]
  \caption{Time in seconds for the training and forecasting of each model on a single core. }
  \label{timing}
  \centering
  \scalebox{0.88}{
   \begin{tabular}{llll}
    \toprule
       Dataset     & Model     &  Training & Forecasting \\
    \midrule
   \multirow{3}{*}[-2pt]{Exchange Rate} &  DLM &  1.9&	37.4 \\
   & TVAR(7) & 0.57 & 124.9\\
   & LSTM(2) & 214.8& 1.4\\
   & LSTM(5) & 222.1& 1.4\\
   & \textbf{Stanza(7)} & 41.8& 669.9 \\
    \bottomrule
  \end{tabular}
  }
\end{table}

\end{document}